%% file: root.tex
\documentclass[letterpaper, 10 pt, conference]{ieeeconf}  

\IEEEoverridecommandlockouts                              

\usepackage{amsmath} 
\usepackage{amssymb}  
\usepackage{mathtools}
\usepackage{xcolor}
\usepackage{graphicx}
\usepackage{lipsum}
\usepackage{algorithm}
\usepackage{algpseudocode}
\usepackage{caption}
\usepackage{url} 
\let\labelindent\relax
\usepackage{enumitem}
\usepackage{siunitx}
\usepackage{kotex}
\usepackage{booktabs} 
\usepackage{color}

\newcommand{\va}{{\bf a}}
\newcommand{\vb}{{\bf b}}

\newcommand{\vf}{{\bf f}}

\newcommand{\vh}{{\bf h}}
\newcommand{\vk}{{\bf k}}

\newcommand{\vo}{{\bf o}}

\newcommand{\vr}{{\bf r}}

\newcommand{\vu}{{\bf u}}

\newcommand{\vw}{{\bf w}}

\newcommand{\vy}{{\bf y}}

\newcommand{\vA}{{\bf A}}
\newcommand{\vB}{{\bf B}}
\newcommand{\vC}{{\bf C}}

\newcommand{\vI}{{\bf I}}

\newcommand{\vtau}{\boldsymbol{\tau}}

\newcommand{\vDelta}{\boldsymbol{\Delta}}
\newcommand\mdoubleplus{\mathbin{+\mkern-10mu+}}
\title{\LARGE \bf
DiSPo: Diffusion-SSM based Policy Learning for\\Coarse-to-Fine Action Discretization
}

\author{Nayoung Oh, Jaehyeong Jang, Moonkyeong Jung, and Daehyung Park\textsuperscript{\textdagger}
\thanks{
The authors are with Korea Advanced Institute of Science and Technology, Korea ({\tt\small \{nyoh, wogud9019, jmk7791, daehyung\}@kaist.ac.kr}). 
{\textsuperscript{\textdagger}}D. Park is the corresponding author.}
\thanks{
This work was supported by the Institute of Information \& communications Technology Planning \& Evaluation(IITP) grant funded by the Korea government(MSIT) (No. RS-2024-00336738, RS-2022-II220311, RS-2024-00509279) and the Technology Innovation Program(RS-2025-25453780) funded by the Ministry of Trade Industry \& Resources(MOTIR, Korea).
}
}

\begin{document}

\makeatletter
    \let\@oldmaketitle\@maketitle
    \renewcommand{\@maketitle}{
    \@oldmaketitle
    \centering
    \vspace{-2.5pt}
    \includegraphics[width=\textwidth]{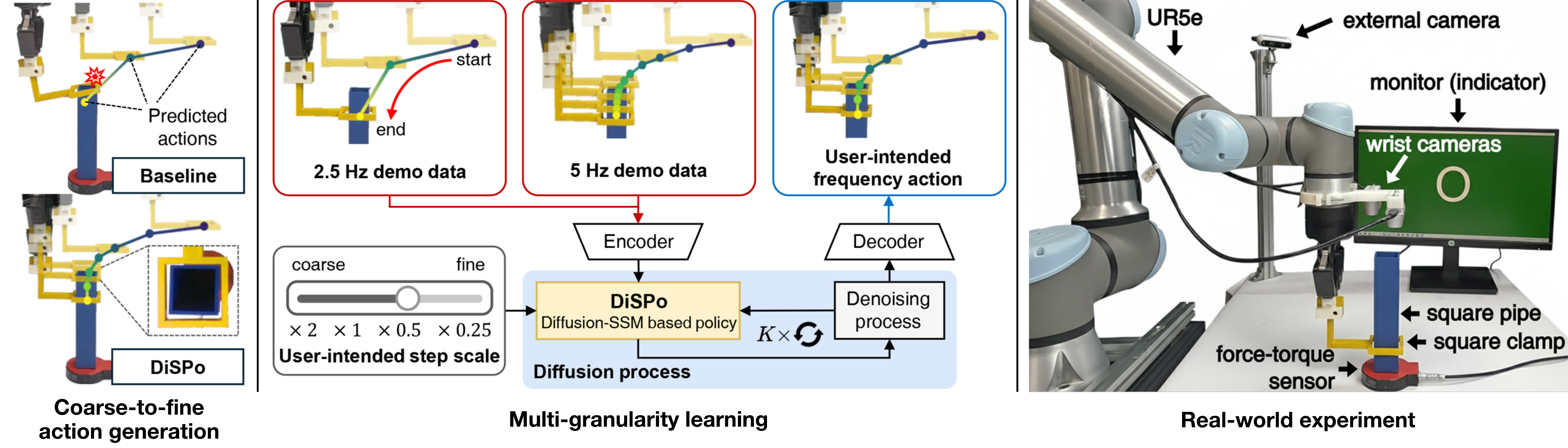}
    \captionof{figure}{Overview of \texttt{DiSPo}: a diffusion-SSM based policy for coarse-to-fine imitation learning. Our proposed \texttt{DiSPo} framework generates fine-grained actions from coarse demonstrations, while the baseline method generates rough actions. Leveraging the representation power of diffusion policy and the flexible discretization capabilities of SSMs, \texttt{DiSPo} learns from multi-granularity demonstrations and generates actions at user-intended granularities. A demonstration of a real-world \textit{clamp passing} task, which requires high precision control to avoid collision.}
    \label{fig:overview}
    \vspace{-9.5pt}
    \addtocounter{figure}{-1}%
    }
\makeatother

\maketitle
\thispagestyle{empty}
\pagestyle{empty}
\addtocounter{figure}{1}

\input{0_abstract}

\input{1_introduction}

\input{3_preliminaries}
\input{4_1_methodology}
\input{4_2_methodology}
\input{5_experimental_setup}
\input{6_evaluation}

\input{7_conclusion}
\input{2_related_work}
\bibliographystyle{ieeetr}
\bibliography{example}
\end{document}

%% file: 0_abstract.tex
\begin{abstract}
We aim to solve the problem of learning user-intended granular skills from multi-granularity demonstrations. Traditional learning-from-demonstration methods typically rely on extensive fine-grained data, interpolation techniques, or dynamics models, which are ineffective at encoding or decoding the diverse granularities inherent in skills. To overcome it, we introduce a novel diffusion-state space model (SSM) based policy (\texttt{DiSPo}) that leverages an SSM, Mamba, to learn from diverse coarse demonstrations and generate multi-scale actions. Our proposed step-scaling mechanism in Mamba is a key innovation, enabling memory-efficient learning, flexible granularity adjustment, and robust representation of multi-granularity data. \texttt{DiSPo} outperforms state-of-the-art baselines on coarse-to-fine benchmarks, achieving up to an $81\%$ improvement in success rates while enhancing inference efficiency by generating inexpensive coarse motions where applicable. We validate \texttt{DiSPo}'s scalability and effectiveness on real-world manipulation scenarios. Code and Videos are available at https://robo-dispo.github.io.
\end{abstract}

%% file: 1_introduction.tex
\section{Introduction}

Researchers have increasingly focused on endowing robots with dexterous, generalizable policies such as human manipulations. These manipulations are often a mixture of coarse-to-fine actions~\cite{johns2021coarse}, we call \textit{multi-granularity} actions. These involve large positioning movements alongside precise maneuvers critical for tasks such as screwing, welding, and insertion. Learning these locally precise behaviors is crucial to task success~\cite{lian2021benchmarking, li2023augmentation, papagiannis2024miles}.

In this context, we aim to solve the problem of generating manipulation skills at multiple levels of granularity through imitation learning (IL), a process we call \textit{multi-granularity learning} as shown in Fig.~\ref{fig:overview}. This requires models to learn from both fine-grained and general coarse demonstrations. Further, the models need to generate precise actions across varying control scales according to user needs, understanding the temporal structure of demonstrations. We term it as \textit{multi-granularity reproduction}. 

Traditional IL methods, such as dynamic movement primitives~\cite{ijspeert2002movement}, learn complex trajectories by adopting dynamics models~\cite{finn2017deep}. These methods allow for frequency adjustments in output, learning from a specific frequency of input trajectories. In the line of research, state space models (SSMs), such as Mamba~\cite{gu2023mamba}, offer memory-efficient, powerful encoding. However, their fixed action representations struggle to capture complex or multi-modal behaviors across diverse task conditions or modalities.

\begin{figure*}[t]
    \centering
    \includegraphics[width=0.9\textwidth]{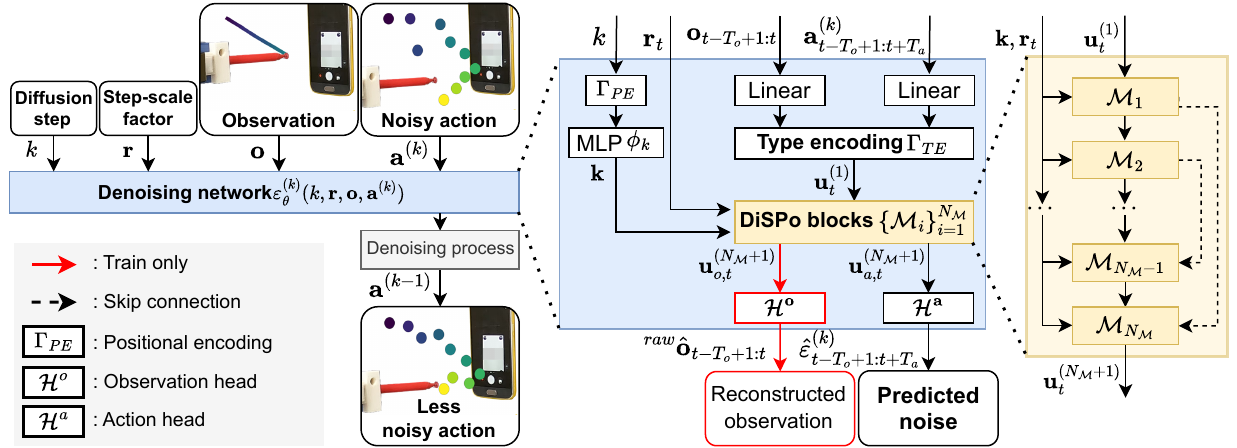}
    \caption{
    Illustration of the \texttt{DiSPo} architecture.
    \texttt{DiSPo} takes diffusion step $k$, step-scale factors $\vr_t$, encoded observations $\mathbf{o}_{t-T_o+1:t}$, and noisy actions $\mathbf{a}^{(k)}_{t-T_o+1:t+T_a}$. The model identifies the noise $\hat{\varepsilon}^{(k)}_{t-T_o+1:t+T_a}$ within the input noisy actions through stacked \texttt{DiSPo} blocks and utilizes the identified noise to generate the less noisy action $\mathbf{a}^{(k-1)}_{t-T_o+1:t+T_a}$.
    } 
    \label{fig:architecture}
\end{figure*}

Alternatively, neural IL methods, such as behavior transformers~\cite{shafiullah2022behavior,lee2024vqbet, gong2024carp} and diffusion-based policies~\cite{janner2022planning, chi2023diffusionpolicy, mishra2023generative, octo_2023}, are increasingly acquiring attention with expressive power and robustness. These approaches are capable of learning from diverse,
high-dimensional multi-modal datasets~\cite{padalkar2023open,brohan2022rt, brohan2023rt, shridhar2023perceiver, wangmimicplay}. However, most approaches learn from a specific frequency of trajectories~\cite{droid} or an unspecified timescale of state-action pairs~\cite{padalkar2023open}, without understanding \textit{multi-granularity}. Further, modeling fine-grained skills typically requires high-frequency demonstrations causing storage and computational overhead.

We propose a novel coarse-to-fine imitation learning algorithm, diffusion-SSM based policy (\texttt{DiSPo}), combining the representation power of diffusion models with the flexible discretization power of SSMs. We particularly adopt a state-of-the-art SSM, Mamba, to enable \texttt{DiSPo} to learn and reproduce trajectories at \textit{multi-granularity} through data-efficient training strategies as shown in Fig.~\ref{fig:overview}. We show that \texttt{DiSPo} is capable of producing varying scales of behavior, not only learning from multiple rates of coarse demonstrations but also modulating the discretization level of trajectories through a granularity predictor online. To the best of our knowledge, this is the first attempt to modulate Mamba's discrete model for fine-grained manipulations. We introduce novel coarse-to-fine IL benchmarks evaluating our method against state-of-the-art visuomotor policy learning methods. The evaluation shows the modulation of step size in \texttt{DiSPo} generates finer movements with expert-like behaviors.

%% file: 3_preliminaries.tex
\section{Preliminaries}
An SSM describes a dynamic system that accepts an input sequence $\vu(t) \in \mathbb{R}$, produces an output sequence $\vy(t) \in \mathbb{R}$, and updates a set of internal states $\vh(t) \in \mathbb{R}^N$, where $N$ denotes the dimensions of the state. The system consists of first-order differential equations, known as state and output equations:
$\dot{\vh}(t) = \vA \vh(t) + \vB \vu(t), \quad \vy(t) = \vC \vh(t)$,
where $\vA\in\mathbb{R}^{N\times N}$, $\vB\in\mathbb{R}^{N\times 1}$, and $\vC\in\mathbb{R}^{1\times N}$ are the state, input, and output parameters, respectively. 

For discrete computations, the SSM transforms the continuous-time system into a discrete-time system, defined over a discrete input sequence $\vu_{1:L} \in \mathbb{R}^{L\times 1}$ and output sequence $\vy_{1:L} \in \mathbb{R}^{L\times 1}$ where $L$ denotes the sequence length. Given a step size $\vDelta\in \mathbb{R}$, the discrete-time system at time step $t$ is 
$
\vh_t = \bar{\vA} \vh_{t-1} + \bar{\vB} \vu_t, \quad \vy_t = \vC \vh_t 
$,
where the discrete parameters are
\begin{align}
\bar{\vA}=\exp{(\Delta \vA)}, \;\; \bar{\vB}=(\Delta\vA)^{-1}(\exp(\Delta\vA)-\vI)\cdot \Delta \vB,
\label{eq:discretization}
\end{align}
following the zero-order hold (ZOH) discretization rule.
In contrast to S4~\cite{s4} with fixed step sizes, Mamba makes parameters $(\vB, \vC, \vDelta)$ as a function of the input $\vu_t$,
\begin{align}
\vB_t = f_B(\vu_t), \: \vC_t=f_C(\vu_t), \: {\vDelta}_t = \text{SoftPlus}(f_{\vDelta} (\vu_t)), 
\label{eq:param_learning}
\end{align}
where $f_B$, $f_C$, and $f_{\vDelta}$ are trainable layers, and $\text{SoftPlus}$ is an activation function. In this work, we extend the SSM to process $D$-dimensional input and output sequences, $\vu_{1:L} \in \mathbb{R}^{L\times D}$ and $\vy_{1:L} \in \mathbb{R}^{L\times D}$. We handle each of the $D$ channels independently, but share the matrix $\vA$ across all channels~\cite{s4, gu2023mamba}.

%% file: 4_1_methodology.tex
\section{Diffusion-SSM based Policy Model}
Our proposed model, \texttt{DiSPo}, generates multi-granularity actions using a Mamba-based diffusion process.
The model progressively denoises an input sequence by predicting the parameter of a discrete SSM (e.g., $\vA$ and $\vB$), conditioned on a user-defined step-scale factor $\vr$. Fig.~\ref{fig:architecture} illustrates the architecture, incorporating a denoising diffusion probabilistic model (DDPM)~\cite{ho2020denoising} with $N_{\mathcal{M}}$ stacked \texttt{DiSPo} blocks $\{\mathcal{M}_i\}_{i=1}^{N_\mathcal{M}}$, where each \texttt{DiSPo} block is a variant of the Mamba block. Inspired by the decoder-only Mamba (\texttt{D-Ma})~\cite{jia2024mail}, we design the architecture to learn denoising networks $\varepsilon_\theta^{(k)}$, parameterized by $\theta$. These networks generate a less noisy action sequence $\va^{(k-1)}$, at the $k$-th denoising step ($k \in [1, \dots, K]$), conditioned on a history of observations $\vo$, noisy actions $\va^{(k)}$, and step-scale factors $\vr$:
\begin{align}
\va^{(k-1)}=\alpha \left( \va^{(k)} - \gamma \varepsilon_\theta^{(k)} (k, \vr, \vo, \va^{(k)}) + \mathcal{N}(0, \sigma^2I )  \right),\!
\label{eq:diffusion_process}
\end{align}
where $\alpha$, $\gamma$, and $\sigma$ are the noise schedule parameters following the DDPM formulation~\cite{ho2020denoising}. 
For notational simplicity, we omit the time index $t$.
Starting from an initial Gaussian noise sample, $\va^{(K)}$, \texttt{DiSPo} recursively applies the denoising process to generate an imitated action sequence. 

A distinct feature of \texttt{DiSPo} is the integration of factors $\vr$ into Mamba, inspired by manual adjustment of rates in time-invariant SSMs~\cite{s4, mondal2024hummuss}. The adjustable factor $\vr$ represents the discretization scale (i.e., temporal granularity) of the sequences in the discrete SSM. This allows \texttt{DiSPo} to learn from multiple rates of demonstrations and to adjust step sizes for discrete SSM parameters. We describe the details below. 

\subsection{Mamba-based denoising process}
Consider an input sequence $\vu_t^{(1)}\in \mathbb{R}^{L \times D}$ in the $k$-th diffusion step and the time step $t$, where $L$ and $D$ are the length and dimension of the input sequence, respectively. Note that, to simplify the notation, we omit $k$ and retain $i$ for the variables defined in the $k$-th step below (e.g. $\vu_{t}^{(k, 1)} = \vu_{t}^{(1)}$). The Mamba-based denoising network predicts the action noise $\hat{\varepsilon}^{(k)}$ by updating the sequences $\vu_t^{(i)}$ with noise-relevant features through the $\{\mathcal{M}_i\}_{i=1}^{N_\mathcal{M}}$ blocks. Then it transforms the action component of the last updated sequence $\vu_{a,t}^{(N_{\mathcal{M}}+1)}$ into the action noise through an output action head $\mathcal{H}^a$,
\begin{align}
\vu_t^{(i+1)} = \mathcal{M}_i(\vk, \vr_t, \vu_t^{(i)}) \quad \text{and}\quad  \hat{\varepsilon}^{(k)} = \mathcal{H}^a( \vu_{a,t}^{(N_{\mathcal{M}}+1)} ),
\end{align}
where $i\in[1, \dots, N_{\mathcal{M}}]$, $\vk\in\mathbb{R}^{D}$ is an embedding for the diffusion step $k$, and $\vr_t\in\mathbb{R}^L$. Each $\mathcal{M}_i$ block processes input sequences with the same size, $\vu_t^{(i)}\in \mathbb{R}^{L \times D}$. The denoising process consists of three parts: input encoding, diffusion process, and noise prediction. 

\begin{figure}[t]
    \centering
    \includegraphics[width=\linewidth]{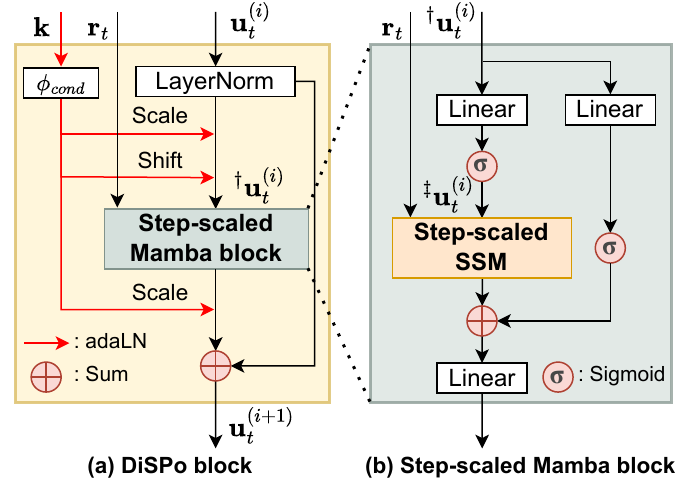}
    \caption{(a) A \texttt{DiSPo} block $\mathcal{M}_i$ refines noise-related features in the type encoded sequence $\mathbf{u}_t^{(i)}$ using adaLN conditioned on the diffusion step embedding $\mathbf{k}$.
    (b) A step-scaled Mamba block takes $\vr_t$ and $^{\dagger}\mathbf{u}_t^{(i)}$.}
    \label{fig:DiSPo_block}
\end{figure}

\noindent\textbf{Input encoding}: The first \texttt{DiSPo} block takes the input sequence $\mathbf{u}_t^{(1)}$, a diffusion step embedding $\vk$, and step-scale factors $\vr_t$ at each $k$-th step.
The input sequence $\mathbf{u}_t^{(1)}$ consists of observation and noisy-action embeddings over lengths $T_o$ and $T_o+T_a$, respectively. We represent it as 
\begin{align}
     \mathbf{u}_t^{(1)}=[&\Gamma_{\text{TE}}(\vf_{o,t-T_o+1}), ... , \Gamma_{\text{TE}}(\vf_{o,t}),  \nonumber \\
     &\Gamma_{\text{TE}}(\vf_{a,t-T_o+1}), ... , \Gamma_{\text{TE}}(\vf_{a,t+T_a}) ],       
\end{align}
where $\vf_{o,t}$ and $\vf_{a,t}$ are observation and action features, respectively. $\Gamma_{TE}: \mathbb{R}^D \rightarrow \mathbb{R}^D$ represent \textit{type encoding}, which injects a learnable vector to the input ($\in \mathbb{R}^D$). Note that $L=2T_o+T_a$.
\begin{figure}[t]
    \centering
    \includegraphics[width=\linewidth]{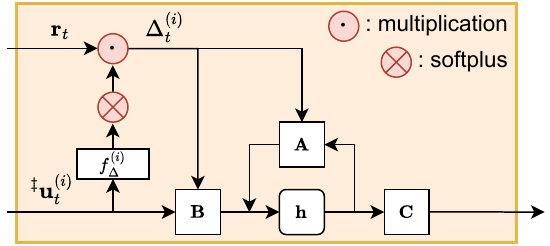}
    \caption{A step-scaled SSM takes input sequence $^{\ddagger}\mathbf{u}_t^{(i)}$ and $\vr_t$ to scale $\Delta_t^{(i)}$, and discretizes the learned SSM parameters using the step sizes.}
    \label{fig:ssm_figure}
\end{figure}
An observation feature $\vf_{o,t}\in\mathbb{R}^D$ is an embedding vector of the observation $\vo_{t}\in \mathbb{R}^{D_{o}}$ preprocessed from raw sensory observations $\prescript{\text{raw}}{}{\vo}_t$ at a timestep $t$. The embedding process is a linear projection by $\vf_{o,t}=\vw_{o} \vo_{t} + \vb_{o}$ with a weight matrix $\vw_o \in \mathbb{R}^{D \times D_{o}}$ and a bias $\vb_o \in \mathbb{R}^{D}$. In this work, we use $\vo_{t}$ as a concatenated vector of an image encoding from ResNet18~\cite{he2016deep} with attentional pooling~\cite{radford2021learning} and a proprioception vector (e.g., end-effector positions) normalized in the range of $[-1, 1]$.

\begin{figure*}
    \centering
    \includegraphics[width=0.9\linewidth]{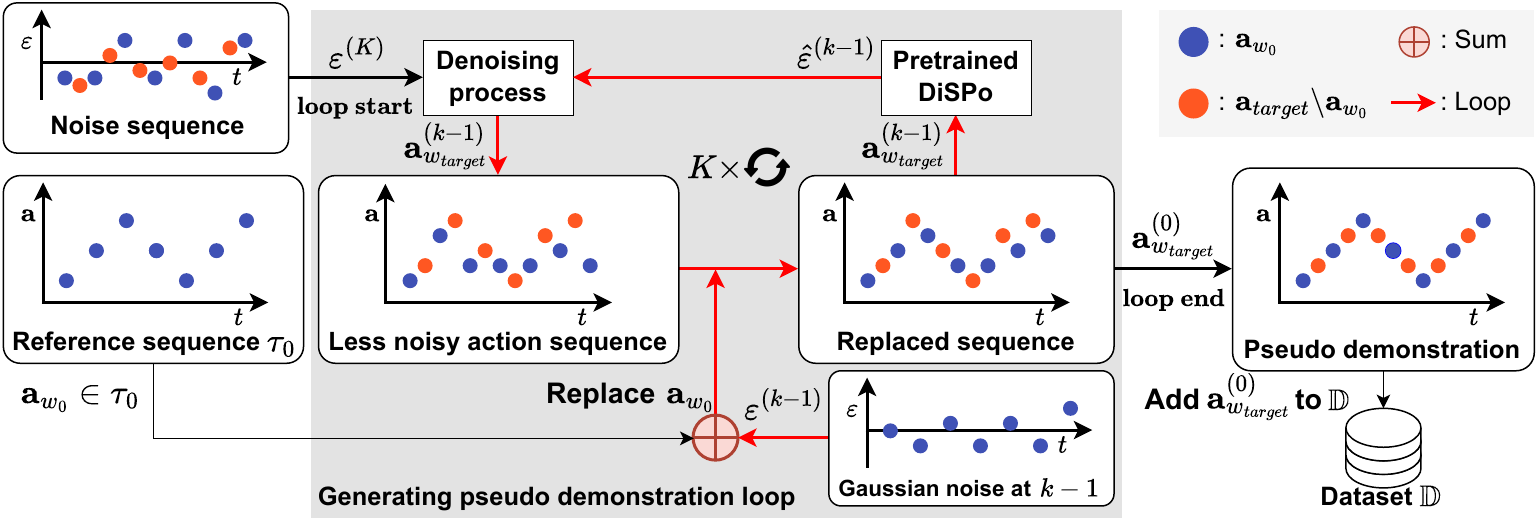}
    \caption{Generating a pseudo demonstration for fine-tuning. Starting from Gaussian noise $\varepsilon^{(K)}$ and a reference sequence $\tau_0$, the model iteratively denoises and replaces $w_0$ frequency actions in the less noisy action sequence with noise added $\mathbf{a}_{w_0}\in\tau_0$. We repeat this process until the model generates a noise-less action sequence at target frequency $\mathbf{a}_{w_{\text{target}}}^{(0)}$, which we refer to as a pseudo demonstration.
    }
    \label{fig:train}
\end{figure*}

An action feature $\vf_{a,t}\in\mathbb{R}^D$ is an embedding vector of the action $\va_t\in \mathbb{R}^{D_a}$, obtained either by normalizing the raw action command $\prescript{\text{raw}}{}{\va}_t$ with noise during training or by denoising the noisy action from the previous diffusion step during inference. The embedding process is a linear projection by $\vf_{a,t} = \vw_a \va_t + \vb_a$ with a weight matrix $\vw_a \in \mathbb{R}^{D \times D_{a}}$ and a bias $\vb_a \in \mathbb{R}^{D}$.
In this work, we use a pose vector as a command, normalizing in the range of $[-1, 1]$. 

Lastly, as a part of input conditions, we embed the diffusion step $k$ into a $D$-dimensional vector $\vk = \phi_k (\Gamma_{\text{PE}}(k))$ by sinusoidal \textit{positional encoding} $\Gamma_{PE}: \mathbb{R} \rightarrow \mathbb{R}^D$ followed by a multi-layer perceptron $\phi_k: \mathbb{R}^D \rightarrow \mathbb{R}^D$.
We provide further details on the step-scale factors $\vr_t$ in Sec.~\ref{sec:multi}.

\noindent\textbf{Diffusion process}: 
At each diffusion step, we update the sequence $\vu_t^{(i)}$ with noise-relevant features through stacked \texttt{DiSPo} blocks $\{\mathcal{M}_i\}_{i=1}^{N_\mathcal{M}}$ with skip connections. Fig.~\ref{fig:DiSPo_block} shows a \texttt{DiSPo} block that is a step-scaled Mamba block with adaptive layer normalization (adaLN)~\cite{peebles2023scalable}, performing dimension-wise scaling and shifting $\vu_t^{(i)}$ into $\prescript{\dagger}{}{\vu}_t^{(i)}$ conditioned on the diffusion step embedding $\vk$. Taking $\prescript{\dagger}{}{\vu}_t^{(i)}$, the step-scaled Mamba block adjusts the parameters of discrete-time SSM, according to user needs, i.e., a vector of step-scale factors $\vr_t\in\mathbb{R}^{L}_{>0}$, and then updates the input sequence 
via the internal step-scaled SSM. In contrast to conventional Mamba blocks, we exclude convolutional layers that limit handling diverse granularity of input sequences due to fixed-size receptive fields.

Fig. \ref{fig:ssm_figure} shows the proposed step-scaled SSM for \textit{multi granularity}. Our SSM predicts the appropriate step size ${\vDelta}_t^{(i)}\in\mathbb{R}_{>0}^{L\times D}$ with respect to
the input sequence $\prescript{\ddagger}{}{\vu}_t^{(i)}$, a non-linear projection of $\prescript{\dagger}{}{\vu}_t^{(i)}$, and the user-intended scales $\vr_t$, 
\begin{align}
{\vDelta}_t^{(i)} = \vr_t \cdot \text{SoftPlus}\left(f_{\vDelta}^{(i)}\left(\prescript{\ddagger}{}{\vu}_t^{(i)}\right)\right),
\end{align}
where $f_{\vDelta}^{(i)}$ is a block-wise trainable linear layer used in Eq.~(\ref{eq:param_learning}). We use $\vDelta_t^{(i)}$ to calculate $\bar{\mathbf{A}}$ and $\bar{\mathbf{B}}$ following Eq.~(\ref{eq:discretization}).

\noindent\textbf{Noise prediction}: After $N_{\mathcal{M}}$ times of feature updates, the action head $\mathcal{H}^a$ predicts the action noise $\hat{\varepsilon}^{(k)}_{t-T_o+1:t+T_a}$ with respect to $\vu_{a,t}^{(N_\mathcal{M}+1)}$ that corresponds to the noisy action input $\va^{(k)}_{t-T_o+1:t+T_a}$ for the $k$-th denoising process. 
We then use the predicted noise to find the denoised action input $\va^{(k-1)}_{t-T_o+1:t+T_a}$ for the next diffusion step $k-1$, following Eq.~(\ref{eq:diffusion_process}) in inference. 

In addition, during training, we enable our model to reconstruct the given raw observation $\prescript{\text{raw}}{}{\vo}_t$ decoding the updated sequence $\vu_{o,t}^{(N_\mathcal{M}+1)}$ through an observation head $\mathcal{H}^o$.
The reconstruction helps the model to keep capturing fine details in observations across layers. Here, the decoder consists of a linear layer for low-dimensional observation and a ResNet18 decoder for image-based observation.

\subsection{Multi-granularity reproduction}\label{sec:multi}
To control the granularity of generated actions, \texttt{DiSPo} takes a vector of step-scale factors,
$\vr_t=[r^o_{t-T_o+1}, ..., r^o_{t}, r^{a}_{t-T_o+1}, ..., r^a_{t}, ..., r^a_{t+T_a}]$, where $r^o_t$ and $r^a_t$ represent the step size scales, we call factors, of the observation $\vo_t$ and the action $\va_t$ relative to those of a reference sequence. We apply identical scales to the observation and past action sequences, such that $\vr^o_{t-T_o+1:t-1} = \vr^a_{t-T_o+1:t-1}$.

To define the reference step size, we use a mode selection approach that chooses the most frequently observed step size in demonstrations. \texttt{DiSPo} then allows for manual selection of the desired step-scale factors. 
For example, we set 
$\vr_t= \mathbf{1}_{t-T_o+1:t} \mdoubleplus \mathbf{1}_{t-T_o+1:t-1}  \mdoubleplus\mathbf{0.5}_{t:t+T_a}$
when we want to achieve twice finer actions and $\vr_t= \mathbf{1}_{t-T_o+1:t} \mdoubleplus \mathbf{1}_{t-T_o+1:t-1}  \mdoubleplus\mathbf{2}_{t:t+T_a}$ for twice coarser actions.
In addition, \texttt{DiSPo} includes a step-scale factor predictor $\phi_{r}$, implemented as an MLP, which predicts a factor $\vr_t$ to accomplish the task given the observation sequence $\vo_{t-T_o+1:t}$.

%% file: 4_2_methodology.tex
\section{Multi-Granularity Policy Learning}
We introduce a \textit{multi-granularity learning} scheme to improve the prediction performance of high-frequency actions that are not available in the demonstration dataset $\mathbb{D}$.
Our scheme consists of two steps: 1) pretraining with sample-rate augmentation and 2) fine-tuning with pseudo actions.

\begin{figure*}[t]
    \centering
    \includegraphics[width=\linewidth]{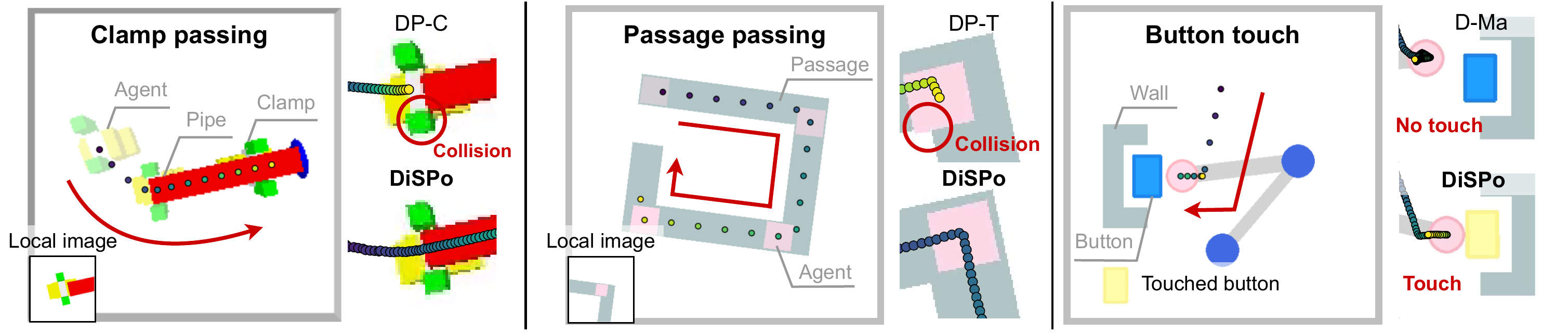}
    \caption{Illustrations of three simulation benchmarks, \textit{clamp passing}, \textit{passage passing}, and \textit{button touch}. Dots denote either demonstrations at \SI{2.5}{\hertz} or predicted actions from \texttt{DiSPo} and baselines. 
    }
    \label{fig:result}
\end{figure*}

In pretraining, to handle various granularities, we first augment the dataset $\mathbb{D}$ with random step-scale factors.
We randomly draw a reference sequence $\tau_0=[(\prescript{\text{raw}}{}{\vo_1}, \prescript{\text{raw}}{}{\va_1}), ..., (\prescript{\text{raw}}{}{\vo_T}, \prescript{\text{raw}}{}{\va_T})]\in \mathbb{D}$ with length $T$ and sample frequency $\omega_0$.
By selecting a frequency $\omega_j\leq \omega_0$, we resample a sequence $\tau_j$ with step-scale factors $\vr_j=\frac{\mathbf{1}_{L}}{(\omega_j/\omega_0)}$ from $\tau_0$. 
Repetition of these enhancements creates the $N_\omega$ number of random frequency sequences: $\vtau = \{ \tau_1, ..., \tau_{N_\omega} \}$. We then introduce a total loss $\mathcal{L}=\mathcal{L}_{MSE}^\varepsilon+ \lambda\cdot \mathcal{L}_{MSE}^o$, 
where $\mathcal{L}_{MSE}^\varepsilon$, $\mathcal{L}_{MSE}^o$, and $\lambda$ are a noise prediction error loss, an observation reconstruction loss, and a weighting coefficient ($\in \mathbb{R}_{> 0}$), respectively. In detail, $\mathcal{L}_{MSE}^\varepsilon$ uses the mean squared error (MSE) to minimize a variational bound on the KL divergence between the true denoising process and that modeled by \texttt{DiSPo}:
\begin{align}
\mathcal{L}_{MSE}^{\varepsilon}=MSE(\varepsilon^{(k)}, \varepsilon_\theta(k, \vr_t, \vo_t, \va_t^{(0)}+\varepsilon^{(k)})).
\end{align}
where $k \in [0, \dots, K]$. Likewise, $\mathcal{L}_{MSE}^o$ is the MSE between an input observation $\prescript{\text{raw}}{}{\vo}_t$ and its reconstruction from the observation head $\mathcal{H}^o$.

In fine-tuning, we co-finetune \texttt{DiSPo} on original and pseudo demonstration dataset to produce high-frequency actions not available in the dataset $\mathbb{D}$.
Fig. \ref{fig:train} shows the process of generating fine-grained pseudo demonstrations. We randomly draw a reference sequence $\tau_0$ with its frequency $w_0$ and generate a fine-grained sequence using the pretrained \texttt{DiSPo} by selecting a target frequency $\omega_{\text{target}} > \omega_0$ with $\vr=\frac{\mathbf{1}_{L}}{(\omega_\text{target}/\omega_0)}$. In practice, starting from Gaussian noise, we perform the diffusion process $K$ times to generate a noise-less action sequence $\va^{(0)}_{1:T'}$, where $T'=T\cdot\omega_{\text{target}}/\omega_0$. However, the pretrained \texttt{DiSPo} is not sufficient to accurately produce high-frequency actions yet. 

To figure it out, we decompose the predicted high-frequency actions $\va^{(k)}_{1:T'}$ into a subset with $\omega_0$ frequency of actions $\va^{(k)}_{w_0}$ and its complement, $\va^{(k)}_{1:T'}\setminus \va^{(k)}_{w_0}$. At each $k$-th denoising process, we replace $\va^{(k)}_{w_0}$ with the demonstration actions in $\tau_0$ with noise, $\va_{1:T}+\varepsilon^{(k)}$. This replacement helps in producing fine-grained pseudo actions that remain close to the demonstrations.
In addition, as \texttt{DiSPo} predicts an action chunk, it produces multiple actions at a timestep.
We aggregate these repeated predictions by weighted averaging, following the temporal ensemble in ACT \cite{act2023}, to obtain the final fine-grained action sequence.
In contrast, the generation of fine-grained observations remains challenging. Thus, we retain the original frequency $\omega_0$ of the observations by setting $r_t^o= 1$ and $r_t^a = \omega_0/\omega_{\text{target}}$ in fine-tuning. We call each outcome sequence a pseudo demonstration. 

We fine-tune \texttt{DiSPo} using both pseudo demonstrations and original demonstrations. In practice, we repeat the generation of pseudo demonstrations and fine-tuning, gradually increasing the target frequency $\omega_{\text{target}}$.
Note that we fine-tune the model with the loss $\mathcal{L}$ corresponding to $\va^{(k)}_{w_0}$ only since the predicted actions $\va^{(k)}_{1:T'}\setminus \va^{(k)}_{w_0}$ are not reliable as original demonstrations. However, sequential prediction with finer step-scale factors helps fine-tuning it as SSM internal state propagates through a sequence. 

%% file: 5_experimental_setup.tex
\section{Experimental Setup}
We conduct quantitative and qualitative evaluations using three simulated benchmarks and two real-world manipulation tasks. The benchmarks statistically assess the ability to generate fine-grained actions from coarse demonstrations. Below, we describe each benchmark in detail.

\noindent\textbf{\textit{Clamp passing}}: A gripper agent (yellow) manipulates a clamp (green) to precisely approach and pass through a 2D pipe (red) without collision, as shown in Fig.~\ref{fig:result} (Left). The raw observation $\prescript{\text{raw}}{}{\vo}_t$ comprises the agent's pose ($\in\mathbb{R}^3$) and two RGB images ($\in\mathbb{Z}^{96\times96\times3}$), one focusing on the agent (Fig.~\ref{fig:result} left local image) and the other capturing the entire scene. The raw action $\prescript{\text{raw}}{}{\va}_t$ is the agent's target pose ($\in\mathbb{R}^3$). We randomize the initial agent pose and vary the pipes’ geometric properties (length and thickness) and spatial pose (position and orientation) using Pybullet~\cite{coumans2019}.

\noindent\textbf{\textit{Passage passing}}: A rectangular agent (pink) precisely maneuvers through a narrow 2D passage (gray) navigating corners without collision, as shown in Fig.~\ref{fig:result} (Middle). The observation $\prescript{\text{raw}}{}{\vo}_t$ includes the agent's pose ($\in \mathbb{R}^2$) and two RGB images as in the \textit{clamp passing} benchmark. The action is the 2D target position aligning the agent with the passage boundary. We randomize the passage's shape, width, and orientation using Pymunk~\cite{Blomqvist2007pymunk} and Pygame~\cite{pygame}. 
\begin{figure}[t]
    \centering
    \includegraphics[width=\linewidth]{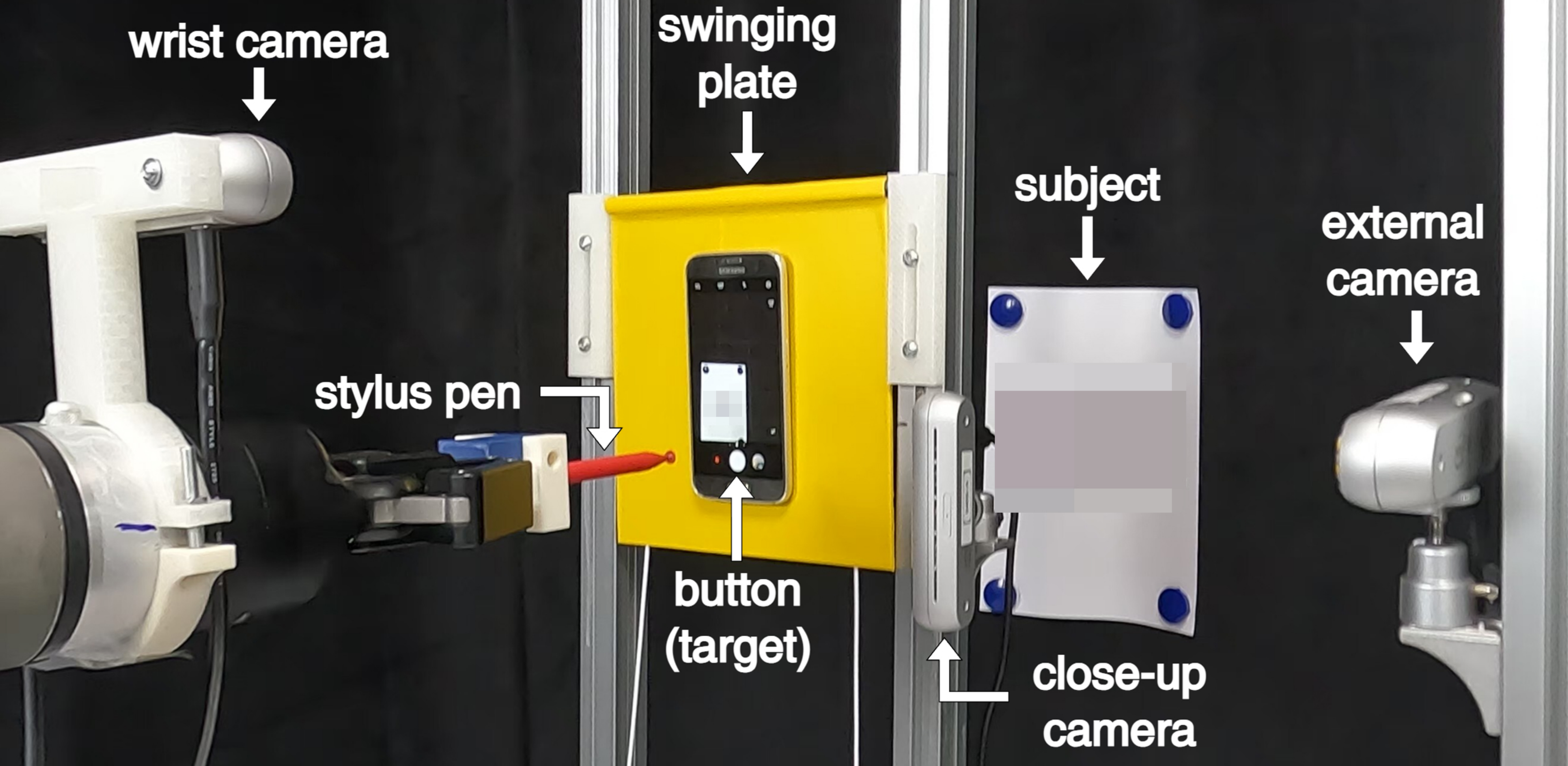}
    \caption{A capture of the real-world \textit{button touch} environment.}
    \label{fig:real_set}
\end{figure}

\noindent\textbf{\textit{Button touch}}: A two-link planar arm precisely touches a button (blue) without causing a collision between the button and wall, as shown in Fig.~\ref{fig:result} (Right). The observation $\prescript{\text{raw}}{}{\vo}_t$ consists of the end-effector position ($\in\mathbb{R}^2$) and an RGB image. The action $\prescript{\text{raw}}{}{\va}_t$ is the desired end-effector position ($\in\mathbb{R}^2$). We randomize the initial arm configuration and button placement, using Pymunk and Pygame. 

For evaluation, we record $90$ demonstrations for each benchmark at frequencies ranging from \SI{2.5}{\hertz} to \SI{20}{\hertz} using the toppra path planning library~\cite{pham2018new}. We then train our method and four baselines on these demonstrations, selecting the best checkpoints based on performance across $50$ randomly sampled validation environments.

\begin{figure*}[t]
    \centering
    \includegraphics[width=0.9\textwidth]{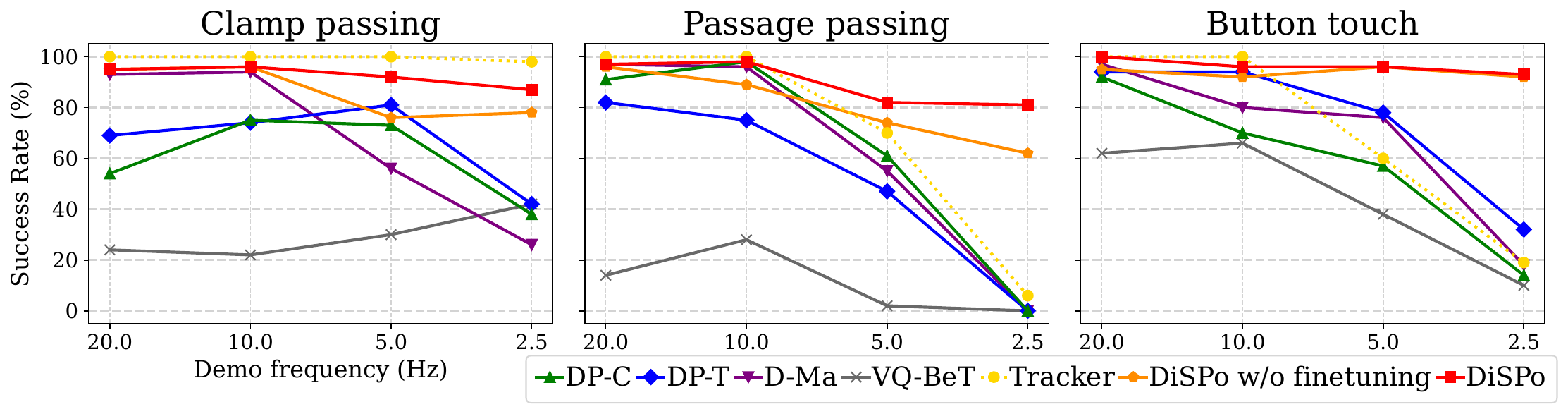}
    \caption{Comparison of task success rates [\%] across four frequencies of demonstrations per simulated benchmark. We train each method with a source frequency ($x$-axis) of demonstrations and test a \SI{20}{\hertz} target frequency of actions in new environments. Note that \texttt{Tracker} is a complexity indicator.}
    \label{fig:coarse-to-fine-benchmarks}
\end{figure*}

We finally evaluate the approaches on $100$ unseen test environments. The four baselines are as follows:
\begin{itemize}[leftmargin=*, noitemsep,topsep=0pt]
    \item DiffusionPolicy-C (\texttt{DP-C}) and DiffusionPolicy-T (\texttt{DP-T})~\cite{chi2023diffusionpolicy}: CNN- and Transformer-based diffusion policies, respectively.
    \item \texttt{D-Ma}~\cite{jia2024mail}: A decoder-only variant of the Mamba-based diffusion model, MaIL.
    \item \texttt{VQ-BeT}~\cite{lee2024vqbet}: A vector-quantized behavior Transformer (BeT) tokenizing continuous actions. 
\end{itemize}
As baselines require fixed step sizes, we linearly interpolate their action sequences. In contrast, \texttt{DiSPo} generates fine-grained actions on demand using user-intended or predicted step-scale factors from the learned predictor $\phi_r$. For comparison, we compute step-scale factors based on the demonstration and required frequency. We also use relative poses as desired actions when advantageous for baselines; baselines adopt relative poses as action representations for the \textit{clamp passing} and \textit{passage passing} tasks except \texttt{DP-C}, known to perform better with absolute positions~\cite{chi2023diffusionpolicy}.
We also report the performance of \texttt{Tracker}, which is not a baseline but a trajectory follower of downsampled ground-truth demonstrations, as a task complexity indicator.

Finally, we demonstrate our method and a baseline, \texttt{D-Ma}, on real-world \textit{clamp passing} as shown in Fig.~\ref{fig:overview} and \textit{button touch} as shown in Fig.~\ref{fig:real_set} tasks using a UR5e manipulator. Unlike the simulated benchmarks, we extend the action space to 3D translation and horizontal rotation ($\in\mathbb{R}^4$) for \textit{clamp passing} and 3D translation for \textit{button touch}. Each task uses three RealSense cameras: two for local views and one for a fixed global view.
We collect $95$ human demonstrations at \SI{2.5}{\hertz} for \textit{clamp passing} and a mixture of \SI{2.5}{\hertz} and \SI{5}{\hertz} for \textit{button touch}, and train both methods on these coarse demonstrations.
We compare two methods in $10$ random environmental setup for each task. For real-time control, we use the denoising diffusion implicit model (DDIM) \cite{ddim}.

%% file: 6_evaluation.tex
\section{Evaluation}
We first evaluate coarse-to-fine IL performance across three benchmarks using demonstrations at various frequencies. 
As shown in Fig.~\ref{fig:coarse-to-fine-benchmarks}, \texttt{DiSPo} consistently achieves the highest success rates of over $81\%$ across all frequencies, whereas baseline performances significantly drop given \SI{2.5}{\hertz} and \SI{5}{\hertz} demonstrations.
For example, baseline methods usually fail at the corner of \textit{passage passing} where \texttt{DiSPo} generates sharp motion as shown in Fig.~\ref{fig:result}.
Occasionally, \texttt{DiSPo} without fine-tuning underperforms compared to baselines at tasks with fine-grained demonstrations, since the tasks are still solvable with low-frequencies demonstrations as the \texttt{Tracker} shows $100\%$ performances.
Nevertheless, our fine-tuning method improves performance by up to $19\%$, with an average gain of $6\%$, without additional data collection. In contrast, the \texttt{Tracker} and baseline performances drop to near zero at \SI{2.5}{\hertz}, failing to reproduce abrupt corner maneuvering. These results highlight \texttt{DiSPo}'s data efficiency and its ability to accurately learn feature spaces from coarse datasets.
\begin{table}[H]
    \centering
      \small
        \caption{Task success rates (\%) under a mixed-frequency (\SI{2.5}{} and \SI{5}{\hertz}) training dataset in the \textit{button touch} task.}
      \begin{tabular}{cccccc}
        \toprule
        Method & \texttt{DP-C} & \texttt{DP-T} & \texttt{D-Ma} & \texttt{VQ-BeT} & \textbf{\texttt{DiSPo}} \\
        \midrule
        Success Rate & 81 & 81 & 83 & 54 & \textbf{93} \\
        \bottomrule
      \end{tabular}
      \label{fig:multi-resol}
\end{table}
\begin{figure}[H]
    \centering
    \includegraphics[width=\linewidth]{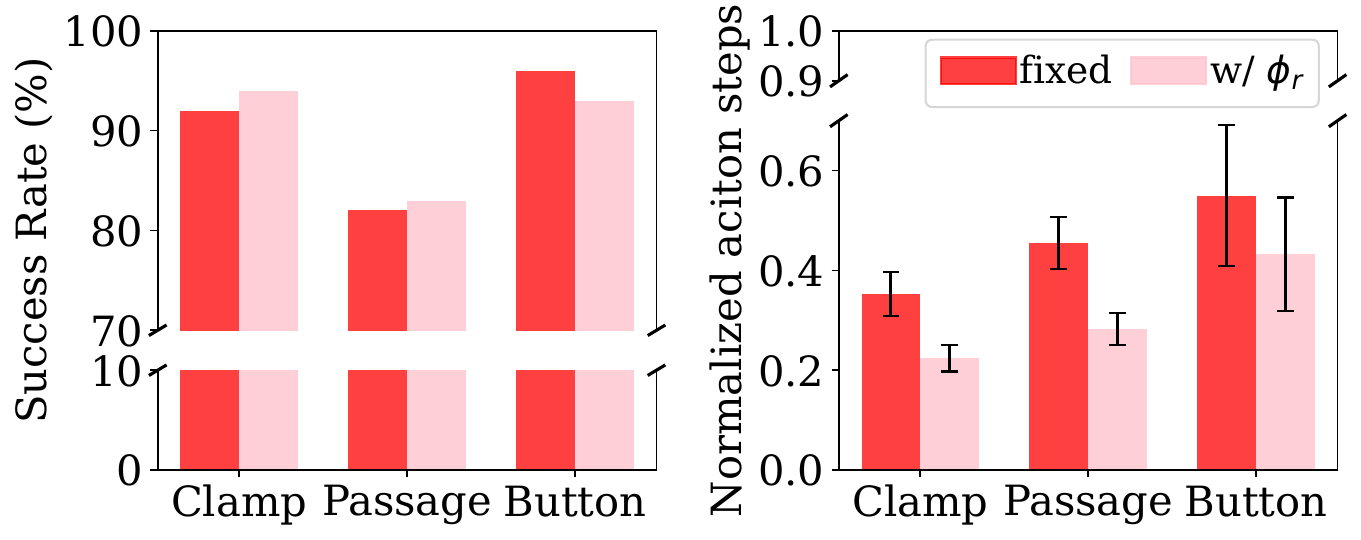}
    \caption{Comparison of \texttt{DiSPo} using fixed versus data-driven step-scaling factors from the predictor $\phi_r$. We normalize the number of action steps taken in successful cases by the maximum step limit allowed in the task.}
    \label{fig:scaling_factor}
\end{figure}

\begin{figure*}[t]   
  \begin{minipage}[t]{0.34\textwidth}
    \centering
    \includegraphics[width=\linewidth]{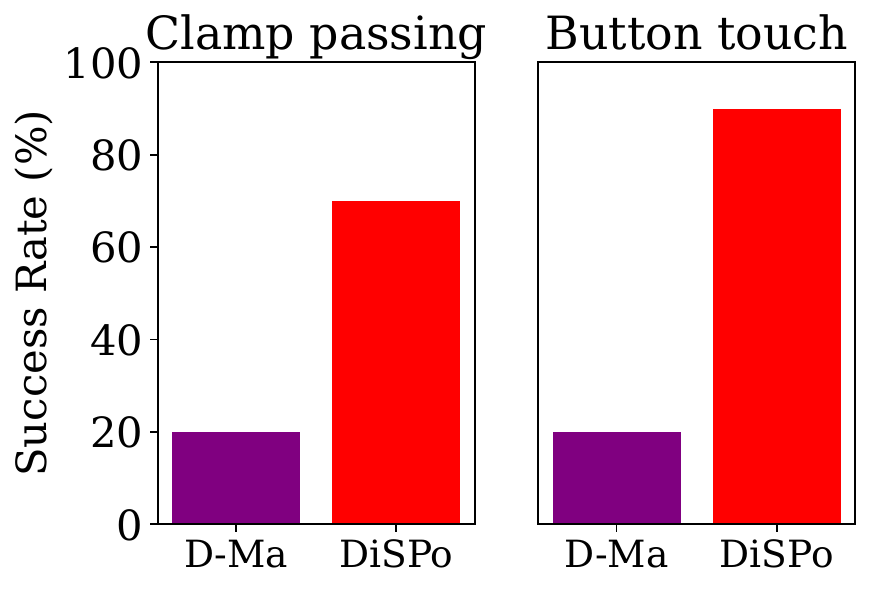}
    \caption{Real-world experiments result, success rates (\%) over $10$ trials}
    \label{tab:real_world}
\end{minipage}
  \hfill
  \begin{minipage}[t]{0.65\textwidth}
    \centering
    \includegraphics[width=\linewidth]{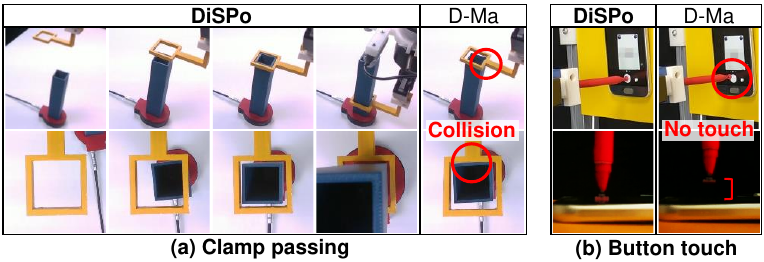}
    \caption{Representative samples showing the UR5e manipulator performing \textit{clamp passing} and \textit{button touch}
    in a real-world environment.}
    \label{fig:real_exp}
  \end{minipage}
\end{figure*}
We also evaluate \textit{multi-granularity learning} by training methods with demonstrations at mixed frequencies, \SI{2.5}{\hertz} and \SI{5}{\hertz}. As shown in Table~\ref{fig:multi-resol}, \texttt{DiSPo} achieves the highest success rate of $93\%$ on the \textit{button touch} task, outperforming all baselines. \texttt{DiSPo} distinguishes sample-wise frequency differences, enabling effective \textit{multi-granularity learning} without performance degradation. In contrast, baselines naively model heterogeneous frequency of state-action pairs, producing actions at an inappropriate speed that cause repetitive small back-and-forth motions near the button.
Using \texttt{DiSPo} trained on \SI{5}{\hertz} demonstrations, we further evaluate the \textit{multi-granularity reproduction} capability of \texttt{DiSPo} applying adaptive step scaling guided by the proposed predictor $\phi_r$. As shown in Fig.~\ref{fig:scaling_factor}, adopted scaling reduces the number of required steps by $32\%$ with only a minor drop in task success rate on the \textit{button touch} task, still significantly outperforming all baselines. These results demonstrate that \texttt{DiSPo} effectively modulates action discretization levels online, producing coarse motions in less critical regions (e.g., free space) to reduce inference overhead while maintaining fine-grained control in critical areas.

Finally, we evaluate \texttt{DiSPo} and \texttt{D-Ma} on a UR5e manipulator in two real-world tasks: \textit{clamp passing} and \textit{button touch}. Fig.~\ref{tab:real_world} shows \texttt{DiSPo} achieves higher success rates than \texttt{D-Ma} in both setups. As shown in Fig.~\ref{fig:real_exp}, \texttt{DiSPo} successfully inserts a square ring clamp with radial clearance \SI{2.5}{\mm} from random initial positions and precisely touches the shutter button by generating fine-grained, collision-free actions.
While \texttt{D-Ma} captures rough motions well, it often causes pipe scratching or stops near the button.

%% file: 7_conclusion.tex
\section{Conclusion}
We proposed \texttt{DiSPo}, a diffusion-SSM based policy that learns and reproduces \textit{multi-granularity} motions by adaptively modulating the action step size with a step-scaling factor. By integrating pseudo-demonstration generation and step-scale prediction, we show that \texttt{DiSPo} reduces storage and computational overhead. On new coarse-to-fine benchmarks, our experimental results demonstrate that \texttt{DiSPo} yields smoother and more accurate motions compared to state-of-the-art methods. Finally, we validate the applicability and superiority of \texttt{DiSPo} through real-world experiments.

%% file: 2_related_work.tex
\section{Related Work}
\noindent\textbf{Learning from demonstration (LfD)}. LfD is a field of methodologies that learn state-action mapping (i.e., policy) from a set of example behaviors~\cite{argall2009survey}.
With the advancements of neural network technology, transformer~\cite{vaswani2017attention} architectures have enabled the encoding of multiple skills with multi-modal inputs~\cite{brohan2022rt, brohan2023rt, shridhar2023perceiver}. To facilitate multi-modal continuous action predictions, researchers have developed categorization-based behavior transformer~\cite{shafiullah2022behavior, lee2024vqbet, gong2024carp} and diffusion-based behavior policy~\cite{janner2022planning, chi2023diffusionpolicy, mishra2023generative, octo_2023}.
Diffusion policy~\cite{chi2023diffusionpolicy}, a representative diffusion model, employs a convolution neural network or transformer backbones for predicting action sequences.

\noindent\textbf{State space models.} Traditional SSMs are to represent a dynamics system with state variables. Adopting the model in deep learning, Gu et al. introduce deep SSMs memorizing continuous-time signals~\cite{gu2020hippo} followed by S4 architecture~\cite{s4} showing fast inference and linear scalability to long sequences.
As a recent variant, Mamba introduces input-dependent adaptability and has gained widespread adoption across diverse research areas by leveraging linear computational complexity and long-range dependency modeling~\cite{gu2023mamba, mamba2, qu2024survey}.
In robotics, Liu et al. integrate the Mamba language model with a vision encoder to construct a reasoning and manipulation model~\cite{liu2024robomamba}.

\noindent\textbf{Diffusion Models with SSMs.} 
Diffusion models typically depend on Transformer backbones for iterative denoising~\cite{peebles2023scalable}. Recent works integrate diffusion models with SSMs to improve efficiency in multiple domains, including time series data \cite{goel2022s, alcarazdiffusion}, vision \cite{yan2024diffusion, hu2024zigma}, and motion generation \cite{zhang2024motion}.
Despite advances in other domains, SSM-driven diffusion remains underexplored in robotics, with prior work focusing mainly on accuracy and efficiency~\cite{jia2024mail, cao2024mamba}. In this work, we instead focus on the discretization feature of SSMs to generate fine-grained motion without additional training.

\noindent\textbf{Coarse-to-fine strategy in robot learning.} 
Most coarse-to-fine manipulation strategies produce actions by refining observations or altering action representations. For observations, researchers either iteratively zoom in on regions of interest to obtain precise data, such as point clouds~\cite{james2022coarse, gervet2023act3d}, or integrate coarse global observations with fine-grained local details~\cite{ling2024articulated,sundaresan2024s}. For actions, they design hierarchical control approaches, alternating between coarse-action generation (e.g., waypoints) and fine-action policies~\cite{johns2021coarse, valassakis2021coarse, belkhale2023hydra}. However, these works primarily aim to improve efficiency when learning from fine demonstrations. Instead, we present a unified policy that handles both coarse and fine actions while enabling learning from coarse demonstrations.